\documentclass[a4paper,11pt]{article}
\usepackage[utf8]{inputenc} 
\usepackage[T1]{fontenc} 
\usepackage{authblk} 
  
\usepackage{hyperref}       
\usepackage{url} 
\usepackage{algorithm} 
\usepackage{amssymb}
\usepackage{amsmath}
\usepackage{algpseudocode} 
\usepackage{bigdelim}
\usepackage{graphicx}
\usepackage{makecell}
\usepackage{tikz}

\begin{document} 
\title{LONG-TERM SERIES FORECASTING WITH QUERY SELECTOR -- EFFICIENT MODEL OF SPARSE ATTENTION}
\author[1,*]{Jacek Klimek}
\author[1]{Jakub Klimek}
\author[1,2]{Witold Kraśkiewicz}
\author[1,2]{\\Mateusz Topolewski}
\affil[1]{MORAI}
\affil[2]{Nicholas Copernicus University}
\affil[*]{Corresponding author Jacek Klimek, jacek.klimek@morai.eu}
\date{}
\maketitle

\section{INTRODUCTION}
Time series  forecasting (TSF) is an old and important area of statistical research with vast practical applications 
to solving real life problems. It has been applied to solving problems arising from many areas of business activities 
such as finance, healthcare, business processes monitoring and others. On the other hand it has been used to model natural 
phenomena such as earthquakes, road traffic ect. At a first glance, it  seems to be a very promising area for application 
of Machine Learning (ML) algorithms, especially those called Deep Learning (DL) methods. However, the situation 
in that matter is not clear yet and even some experts of a more traditional  approach to TSF are expressing 
their general doubts whether DL methods are most appropriate for TSF 
(\cite{elsayed2021really}). 
We believe that it is not the case and if remaining obstacles hindering progress in this area are overcome, 
then DL methods will prevail - similar to what happened in the Natural Language Processing  (NLP) area. 
We think that the main remaining obstacles lay in building right DL models for TSF and finding software implementations 
able to deal with the heavy burden of computations associated with modeling TSF.

\section{RELATED WORK}
As we mentioned above we define work to be related to our research when it can be placed in one of the following two categories. First category contains works that are concerned with finding appropriate general algorithmic approaches to modeling TSF using DL methods. The second category contains works that are dealing with computational efficiency of particular DL models. In the first category, we would like to mention works that use older neural networks architecture such as RNN, CNN, LSTM.
 (\cite{wen2018multihorizon},
 \cite{yu2019longterm}). 
However, since the emergence of Transformer models
(\cite{AttentionAyN})  
 and their overwhelming success in ML research areas stretching
from NLP to Computer Vision we observe the same phenomenon happening in the area of TSF.
In our opinion the most interesting 
papers describing application of Transformer architecture to solving TSF problems are
\cite{zhou2021informer}, \cite{li2020enhancing}, \cite{adversalialTransf}. Most approaches to modeling TSF are based on 
ideas coming from NLP i.e. sequence to sequence framework 
(\cite{sutskever2014sequence}, 
\cite{wen2018multihorizon}, 
\cite{yu2019longterm}). 
However, there are some other approaches utilizing  different ideas - for instance 
(\cite{oreshkin2020nbeats}).

In the second category of works the most interesting, from our point of view, are papers dealing with the most 
fundamental Transformer inefficiency i.e. computational cost of attention mechanism. The most common solution 
to this problem comes down to introducing some sparsity to the attention matrix. One can mention works: 
\cite{beltagy2020longformer}, 
\cite{child2019generating}. 
Additionally, there are some more technical enhancements that can be used to deal with this computational inefficiency 
by fully utilizing hardware capacity - DeepSeed
(\cite{ren2021zerooffload})
or 
FairScale (https://pypi.org/project/fairscale/)

 \section{BACKGROUND (Time series and forecasting)}
In statistics, time series is a collection of random variables $X=(X_t)$ indexed in time order. 
Usually time indeces are equally spaced.
In practice we  identify the time series with its sample path -- a sequence of reals or real vectors of the same dimension.
Time series forecasting is, roughly speaking, a problem of predicting future values of $X$ based 
on previously observed values.
Statistical approach to this problem is to find a conditional distribution of $(X_{n+1}, X_{n+2},\dots, X_{n+k})$ given 
$(X_1, X_2,\dots, X_n)$ and then predict future values using this distribution. There are many ways to solve this problem. 
If $X$ is sufficiently regular, one can try to fit one of ARIMA-type models (involving autoregressive and moving average 
type of dependence). There is a wide literature concerning properties and methods of fitting the ARIMA-type models, 
see e.g. \cite{Box, ARIMA_JAM}. Another very popular (especially in econometrics) models of dependence are so-called ARCH 
(autoregressive conditional heteroscedasticity) models introduced by Robert F. Engle in \cite{engle1982}. The 
main idea of ARCH models is that the variance of $X_t$ is random and depends on the past of $X$. 
More about properties and methods of fitting of ARCH models the reader can find in \cite{Hassler}.
Unfortunately, in most applications those very well described models are not achieving the best results.
 In general situation, one can try 
to decompose the series into 3 components: trend, seasonality and noise, i.e. rewrite the series X as a sum:
\begin{equation}\label{eq.tsn}
                                                                    X=T+S+N,
\end{equation}																	
where $T$ depends linearly on time (with some nonperiodic changes, usually related to „business cycle”), 
$S$ is periodic and $N$ is a white noise (usually Gaussian). Such decomposition provides a useful abstract model for thinking 
about time series generally. For details concerning construction and usage of the decomposition (1) we defer the reader to wide literature devoted to this problem (e.g. \cite{Hyndman}).
In machine learning, the time series forecasting problem is solved in many different ways. 
Some of them involve decomposition (1) and deep learning methods (see e.g. \cite{oreshkin2020nbeats, cleveland90}) and some use only classical ML methods 
like GBRT (e.g. \cite{elsayed2021really}). There is a wide literature devoted to sequence-to-sequence models for TSF (see e.g. \cite{kondo2019sequence, wu2020connecting, lai2018modeling}). 
Nowadays, the Transformer-based models are very popular in deep learning, in particular in sequence-to-sequence models. 
The main advantage of Transformer is that it is able to learn and model all dependencies between input data. 
This property is invaluable in NLP problems, but in time series case the decomposition (1) suggests that the number of 
dependencies is not so large, especially when $n$ is much larger then the length of the period of $S$. Therefore it is 
reasonable to consider sparse versions of attention. 
There are many methods of construction of the sparse version of attention and a lot of them involve some random effects 
(see e.g. \cite{zhou2021informer, child2019generating}). We believe that for validation and comparison reasons it is valuable to use a deterministic method to build 
a sparse version of attention. That is why we propose the method described in section 3. It is a result of a number of 
experiments with different ideas of sparse attention mechanisms in TSF.

\section{METHODOLOGY}
 As we mentioned above, the problem of Time Series Forecasting gets more and more computationally challenging along 
 with increasing length of input data and forecasting range.  Thus, the computational hindrances are an inherent problem 
 in the area of TSF research.  In order to deal with them we researched several methods aiming to reduce this complexity 
 by approximating the full attention matrix by some sparse matrices.  We find that  the following approaches to this problem 
 are the most interesting –- LogSparse \cite{li2020enhancing}, Reformer \cite{kitaev2020reformer} and 
 Informer \cite{zhou2021informer}.  
 Moreover, since Informer reports the best results 
 in comparison to the other mentioned systems, we focused our analysis on this algorithm.  
 The clue of Informer approach seems to be contained in so called Probability Sparse Attention.  
 It relies on a clever choice of indexes that will be used in computing a matrix  that approximates usual attention matrix 
 in vanilla Transformers.  In the process of analyzing and experimenting with this  methodology we come to an impression that, 
 in the case of long sequence TSF, Transformer based systems can converge to optimum using the whole variety of probabilistic 
 choices of indexes  for approximations of attention matrices.  The one of main disadvantages of using probabilistic methods 
 is the difficulty of comparing different experiments with changing parameters.  Thus, we propose a simple but very effective, 
 completely deterministic method for computing sparse approximation of attention matrix which yields excellent results 
 on various time series datasets.  
 
 \begin{algorithm}[h!]\label{algor}
	\caption{Query Selector} 
	\begin{algorithmic}[1]
    \Require $ K \in \mathbb{R}^{L \times D},\ Q \in \mathbb{R}^{L \times D},\ V \in \mathbb{R}^{L \times E},\ 0<f<1 $
    \State set $ \ell = \lfloor (1 - f) \times L\rfloor $
    \State set $ \mathcal{I} = \{ X \in \mathcal{P}(\{ 1,2,...,L \}) : |X| = \ell \} $
    \State set $$ \hat{K} = \frac1{\ell}\cdot( \max\limits_{I \in \mathcal I } \sum\limits_{i \in I} K_{i,1},\ \max\limits_{I \in \mathcal I} \sum\limits_{i \in I} K_{i,2},\ \dots,\ \max\limits_{I \in \mathcal I } \sum\limits_{i \in I} K_{i,D} )  $$
    \State set $ S  = \hat{K} \times Q^T $
	
    \For{\texttt{$i$ in $1\dots\ell$}}\Comment{find indeces of $\ell$ greatest coordinates in $S$}
		\State $c_i=\arg\max S$
		\State $S_{c_i}=-\infty$
	\EndFor
    \State set \[ \hat{Q} = \begin{bmatrix} 
      q_{c_1,1} & q_{c_1,2} & \dots & q_{c_1,D}\\
	  q_{c_2,1} & q_{c_2,2} & \dots & q_{c_2,D}\\
      \vdots & \vdots &\ddots & \vdots\\
      q_{c_\ell,1} & q_{c_\ell,2}& \dots       & q_{c_\ell,D} 
      \end{bmatrix}
    \]
    \State set $ \hat{QK} = \hat{Q} \times K^T$
    \State set $ A = \operatorname{softmax}(\hat{QK} / \sqrt{D}) \times V $
    \State set $$ v = \frac1L\cdot (\sum_{i = 1}^{L}V_{i,1},\ \sum_{i = 1}^{L}V_{i,2},\ ...,\ \sum_{i = 1}^{L}V_{i,E} ) $$
     \State set \[ \hat{A} = \left.\left[ \begin{array}{ {c} }
    v      \\
    v    \\
    \vdots \\
    v  
    \end{array} \right]\   \right\}\hbox{$ L $ rows } \]

    \For{\texttt{$i$ in $1\dots\ell$}}
      \For{\texttt{$j$ in $1\dots E$}}
          \State \texttt{$\hat{A}_{c_i,j} = A_{i,j}$}
        \EndFor
      \EndFor
	  \State \Return $\hat{A}$
  \end{algorithmic} 
\end{algorithm}
 
Let us recall the basic notions of Transformer architecture and set necessary notation. 
Given the input representation $X$, the attention matrix can be calculated as follows.  
First, we compute the query, key and value matrices for each attention head through linear projections, i. e. , $Q = XW_Q$, 
$K =XW_K$ and $V=XW_V$, where $Q$, $K$ and $V$ denote query,  key and value matrices respectively, $W_Q$, $W_K$ and $W_V$ 
are linear projections.  
Then, the attention matrix is computed by a scaled dot-product operation:
$$\operatorname{Attention}(X)=\operatorname{softmax}(QK^T/\sqrt{d})V,$$
were $d$ denotes the hidden dimension.  In order to incorporate sequential information into the model, we utilize (as usual)  
a positional encoding to the input representation.  Here, we follow the original implementation of absolute positional 
encoding (\cite{AttentionAyN}), it is added to the input representation $X$ directly.  
It is worth mentioning here that Informer  has its original proposition for positional encoding. 

In our approach, we are mostly concerned with lowering computational complexity of the algorithm. We called our algorithm 
Query Selector because it’s main insight relies on proper selection of some number of  queries that we will use in our 
computations.  Additionally, we would like to lower the amount of memory needed to store the whole $QK^T$.  
To this end, we construct a matrix $\hat{A}$ which will play the analogous role to role of attention matrix in the 
Transformer algorithm. 

  Another insight for our algorithm comes from the idea of dropout that is successfully employed in many different 
  ML algorithms.  As in dropout we choose a number $f$, $0 < f < 1$, that is one of the hyperparameters of our algorithm.  
  Then, the construction of the matrix $\hat{A}$ is as follows.  We choose $\ell =\lfloor (1 - f) \times L\rfloor $  queries 
  that give the biggest scalar products with keys.  Next, we  replace the usual matrix $K$ with a column-constant matrix 
  $\overline{K}$ of elements equal to the mean value of $\ell$ greatest elements in the column of $K$.   
  Next, we construct $\overline{Q}$ by choosing $\ell$ rows of  matrix $Q$ with indices equal to indices of $\ell$ columns 
  of  $\overline{K}$ with highest common value of given column and remaining rows are set to zero.   
  Matrix  $\hat{A}$ is given by formula
$$\hat{A}  = \operatorname{softmax}(\overline{Q}\cdot \overline{K}^T/\sqrt{D})\cdot V.$$  
It is worth mentioning that this formula implies that the rows of $\hat{A}$ corresponding to zero rows of $\overline{Q}$ 
can be computed by taking the arithmetic means of columns of $V$. 
The details of the above described algorithm are given in the pseudo code on page \pageref{algor}.

\section{EXPERIMENTS}
\subsection{Datasets}
In our experiments we use the  ETT (Electricity Transformer Temperature) dataset. 
In this dataset each data point consists of 8 variables, including the date of the point, the oil temperature and 6 
other parameters measured at electric transformer stations in China.  As we can see from this description, 
ETT dataset allows not only for univariate TSF but  for multivariate as well. ETT dataset consists of three different 
datasets: ETT - small, ETT - large, ETT-full. So far, the only publicly available dataset is ETT - small so we perform 
our experiments using this set of data only. Moreover,  Informer \cite{zhou2021informer} reports the State of the Art (SOTA) results on this 
dataset, so in order to be able to compare our methodology to theirs, we closely follow their experimental setup.  
ETT contains data collected from two regions of a province of China. Datasets  are  named ETT-small-m1 and ETT-small-m2 
and m denotes the fact that data has been measured every minute for 2 years. This gives 70,080 data records altogether. 
Additionally, the ETT dataset contains ETT-small-h1 and ETT-small-h2 subsets with hourly measurements of parameters of 
the same type. More details and access to data can be found at https://github.com/zhouhaoyi/ETDataset.

{\subsection{Experimental Details}

\subsubsection{Baselines}
As we already mentioned, Informer reports the SOTA results on ETT dataset and carried out comparison with other 
important TSF algorithms such that 
\cite{ariyo2014stock}, Prophet \cite{taylor2018forecasting}, LSTMa \cite{bahdanau2014neural}, 
LSTnet \cite{lai2018modeling},  DeepAR \cite{flunkert2017deepar},
Reformer \cite{kitaev2020reformer}, 
and LogSparse self-attention \cite{li2020enhancing}.  
So we compare our results on this dataset with only Informer’s results and our implementation of full attention 
Transformer based on DeepSpeed framework. We use two evaluation metrics, 
$\operatorname{MSE} = \frac1n\sum{(y_i-\hat{y_i})^2}$ and $\operatorname{MAE} = \frac1n\sum|y_i-\hat{y_i}|$.
For multivariate prediction, we average metrics over coordinates.  They are the most popular measures of  prediction accuracy.
Unfortunately they are dependent on the scale of the data.   While scale independent measures 
(like MAPE, sMAPE and others) gain their popularity in literature (see e.g. \cite{Wallstrom}), 
 we avoid scale dependency by normalization of data.

The source code is available at  https://github.com/moraieu/query-selector.
All the models were trained and tested on a 2 Nvidia V100
16 GB GPU.

\subsubsection{Hyper-parameter tuning}   

\begin{table}[t]
\begin{center}
\begin{tabular}{l}
\hline
hidden size: 96, 128, 144, 256, 312, 378, 384, 400, 512\\
embedding size: 16, 18, 24, 32, 48, 64, 96, 112, 128\\
encoder layers: 1, 2, 3, 4\\
decoder layers: 1, 2, 3\\
heads: 2, 3, 4, 5, 6\\
batch size: 24, 32, 48, 64, 96, 100, 128, 144\\
dropout: 0, 0.05, 0.1, 0.15\\
iterations: 1, 2, 3, 4, 5, 6, 7\\
factor $f$:  0.1, 0.2, 0.5, 0.7 ,0.75, 0.8, 0.85, 0.9, 0.95\\
\hline
\end{tabular}
\end{center}
\caption{Hyper-parameters and their ranges used in search}
\end{table}

We conduct a search over the hyper-parameters.
Tuned parameters include number of encoder and decoder layers, 
number of heads, batch size, hidden dimension size, value dimension size, iterations number and dropout. 
For cases using Query Selector we tested different values of query selector factor $f$.
Ranges of search are given in Table 1 and the final values can be found on github 
(https://github.com/moraieu/query-selector/tree/master/settings). 

\begin{figure}
\begin{tikzpicture}
\newlength\hu
\setlength\hu{20cm}
\newlength\vu
\setlength\vu{30cm}
\draw (.475\hu,.20\vu) rectangle (.925\hu,.34\vu);
	\filldraw (.5\hu,.20\vu) circle (1pt) node[below] {$0.5$}; 
	\filldraw (.6\hu,.20\vu) circle (1pt) node[below] {$0.6$}; 
	\filldraw (.7\hu,.20\vu) circle (1pt) node[below] {$0.7$}; 
	\filldraw (.8\hu,.20\vu) circle (1pt) node[below] {$0.8$}; 
	\filldraw (.9\hu,.20\vu) circle (1pt) node[below] {$0.9$};
	
\filldraw (.475\hu,.22\vu) circle (1pt) node[left] {$0.22$};
\filldraw (.475\hu,.24\vu) circle (1pt) node[left] {$0.24$};
\filldraw (.475\hu,.26\vu) circle (1pt) node[left] {$0.26$};
\filldraw (.475\hu,.28\vu) circle (1pt) node[left] {$0.28$};
\filldraw (.475\hu,.30\vu) circle (1pt) node[left] {$0.30$};
\filldraw (.475\hu,.32\vu) circle (1pt) node[left] {$0.32$};	

\draw[violet] (.5\hu, 0.284692\vu)--(.6\hu,  0.31277\vu)--(.7\hu, 0.313502\vu)--(.8\hu, 0.310082\vu)--(.9\hu, 0.328614\vu) (.95\hu, .27\vu)++(1ex,-2ex)--++(.7cm,0)  node[right] {240};
\draw[blue] (.5\hu, 0.260276\vu)--(.6\hu,  0.280124\vu)--(.7\hu, 0.264966\vu)--(.8\hu, 0.285304\vu)--(.9\hu, 0.27322\vu) (.95\hu, .27\vu)++(1ex,-4ex)--++(.7cm,0)  node[right] {360};
\draw[green!80!black] (.5\hu, 0.258252\vu)--(.6\hu,  0.308888\vu)--(.7\hu, 0.306396\vu)--(.8\hu, 0.302782\vu)--(.9\hu, 0.26084\vu)(.95\hu, .27\vu)++(1ex,-6ex)--++(.7cm,0)  node[right] {480};
\draw[brown] (.5\hu, 0.24436\vu)--(.6\hu,  0.261086\vu)--(.7\hu, 0.295898\vu)--(.8\hu, 0.301904\vu)--(.9\hu, 0.232348\vu)(.95\hu, .27\vu)++(1ex,-8ex)--++(.7cm,0)  node[right] {600};
\draw[orange] (.5\hu, 0.245606\vu)--(.6\hu,  0.286448\vu)--(.7\hu, 0.272534\vu)--(.8\hu, 0.263136\vu)--(.9\hu, 0.28528\vu)(.95\hu, .27\vu)++(1ex,-10ex)--++(.7cm,0)  node[right] {720};
\draw[red] (.5\hu, 0.217176\vu)--(.6\hu,  0.276712\vu)--(.7\hu, 0.213624\vu)--(.8\hu, 0.216284\vu)--(.9\hu, 0.288624\vu)(.95\hu, .27\vu)++(1ex,-12ex)--++(.7cm,0)  node[right] {840};
\draw (.7\hu,.175\vu) node {factor $f$};
\draw (.4175\hu, .27\vu) node {\rotatebox[origin=c]{90}{MSE}};
\draw (.95\hu, .27\vu) node[right] {input length};
\end{tikzpicture}
\caption{Influence on MSE of Query Selector factor $f$  for different input lengths.}
\end{figure}
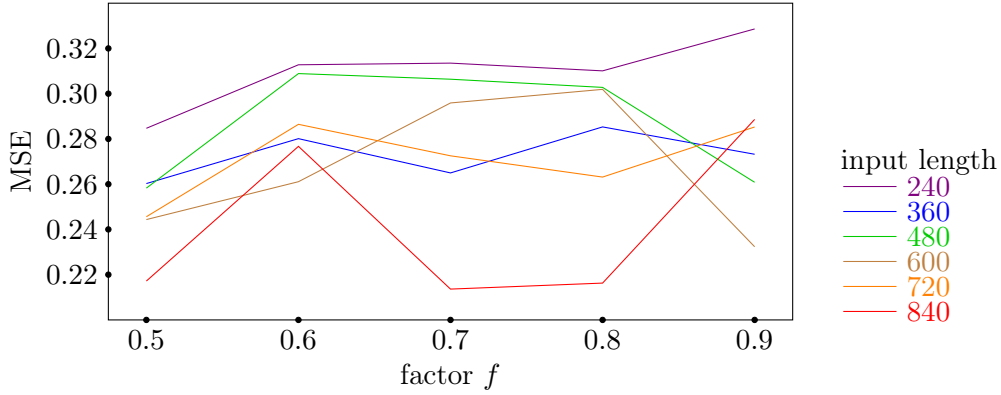

\break 

\subsubsection{Results summery}

\begin{table}[b!]
\begin{tabular}{|*{9}{c|}}
\hline
{\small Data} &	\makecell{\small Pre-\\diction\\ len} &	\makecell{\small Informer\\ MSE} &	\makecell{\small Informer\\ MAE} &	\makecell{Trans\\ former\\ MSE}
 &	\makecell{\small Trans\\ former\\ MAE} &	\makecell{\small Query\\ Selector\\ MSE} &
	\makecell{\small Query\\ Selector\\ MAE} &	\makecell{\small MSE\\ ratio}\\
\hline
\multirowcell{5}{\rotatebox[origin=c]{90}{ETTh1}}
 &	24 &	0.0980 &	0.2470 &	0.0548 &	0.1830 &	\bf 0.0436 &	\bf 0.1616 &	0.445\\
 &	48 &	0.1580 &	0.3190 &	0.0740 &	0.2144 &	\bf 0.0721 &	\bf 0.2118 &	0.456\\
 &	168 & 	0.1830 &	0.3460 &	0.1049 &	0.2539 &	\bf 0.0935 &	\bf 0.2371 &	0.511\\
 &	336 & 	0.2220 &	0.3870 &	0.1541 &	0.3201 &	\bf 0.1267 &	\bf 0.2844 &	0.571\\
 &	720 &	0.2690 &	0.4350 &	0.2501 &	0.4213 &	\bf 0.2136 &	\bf 0.3730 &	0.794\\
 \hline
\multirowcell{5}{\rotatebox[origin=c]{90}{ETTh2}}
 &	24 &	0.0930 &	0.2400 &	0.0999 &	0.2479 &	\bf 0.0843 &	\bf 0.2239 &	0.906\\
 &	48 &	0.1550 &	0.3140 &	0.1218 &	0.2763 &	\bf 0.1117 &	\bf 0.2622 &	0.721\\
 &	168 &	0.2320 &	0.3890 &	0.1974 &	0.3547 &	\bf 0.1753 &	\bf 0.3322 &	0.756\\
 &	336 & 	0.2630 &	0.4170 &	0.2191 &	0.3805 &	\bf 0.2088 &	\bf 0.3710 &	0.794\\
 &	720 &	0.2770 &	0.4310 &	0.2853 &	0.4340 &	\bf 0.2585 &	\bf 0.4130 &	0.933\\
\hline
\multirowcell{5}{\rotatebox[origin=c]{90}{ETTm1}}
 &	24 &	0.0300 &	0.1370 &	0.0143 &	0.0894 &	\bf 0.0139 &	\bf 0.0870 &	0.463\\
 &	48 &	0.0690 &	0.2030 &	\bf 0.0328 &	\bf 0.1388 &	0.0342 &	0.1408 &	0.475\\
 &  96 &    0.1940 &    0.2030 &    \bf 0.0695 &    \bf 0.2085 &    0.0702 &    0.2100 &    0.358\\
 &	288 &	0.4010 &	0.5540 &	\bf 0.1316 &	\bf 0.2948 &	0.1548 &	0.3240 &	0.328\\
 &	672 &	0.5120 &	0.6440 &	\bf 0.1728 &	0.3437 &	0.1735 &	\bf 0.3427 &	0.338\\
 \hline
\end{tabular}
\caption{Forecasting results for univariate  time-series.}
\end{table}

The results for Query Selector are summarized in Table 2 and Table 3. They are compared to the results 
for Informer and for Transformer published in \cite{zhou2021informer}. In prevail number of cases, prediction error of 
Query Selector is smaller or comparable to the error of two other models. 
In the last column of both tables, ratio of MSE for Query Selector to MSE for Informer is given.

\begin{table}[t!]
\begin{tabular}{|*{9}{c|}}
\hline
{\small Data} &	\makecell{\small Pre-\\diction\\ len} &	\makecell{\small Informer\\ MSE} &	\makecell{\small Informer\\ MAE} &	\makecell{Trans\\ former\\ MSE}
 &	\makecell{\small Trans\\ former\\ MAE} &	\makecell{\small Query\\ Selector\\ MSE} &
	\makecell{\small Query\\ Selector\\ MAE} &	\makecell{\small MSE\\ ratio}\\
\hline
\multirowcell{5}{\rotatebox[origin=c]{90}{ETTh1}}
  &	24 &	0.5770 &	0.5490 &	0.4496 &	0.4788 &	\bf 0.4226 &	\bf 0.4627 &	0.732\\
  &	48 &	0.6850 &	0.6250 &	0.4668 &	0.4968 &	\bf 0.4581 &	\bf 0.4878 &	0.669\\
  &	168 &	0.9310 &	0.7520 &	0.7146 &	0.6325 &	\bf 0.6835 &	\bf 0.6088 &	0.734\\
  &	336 &	1.1280 &	0.8730 &	\bf 0.8321 &	0.7041 &	0.8503 &	\bf 0.7039 &	0.738\\
  &	720 &	1.2150 &	0.8960 &	\bf 1.1080 &	\bf 0.8399 &	1.1150 &	0.8428 &	0.912\\
\hline
\multirowcell{5}{\rotatebox[origin=c]{90}{ETTh2}}
  &	24 &	0.7200 &	0.6650 &	0.4237 &	0.5013 &	\bf 0.4124 &	\bf 0.4864 &	0.573\\
  &	48 &	1.4570 &	1.0010 &	1.5220 &	0.9488 &	\bf 1.4074 &	\bf 0.9317 &	0.966\\
  &	168 &	3.4890 &	1.5150 &	\bf 1.6225 &	\bf 0.9726 &	1.7385 &	1.0125 &	0.465\\
  &	336 & 	2.7230 &	1.3400 &	2.6617 &	\bf 1.2189 &	\bf 2.3168 &	1.1859 &	0.851\\
  &	720 &	3.4670 &	1.4730 &	3.1805 &	1.3668 &	\bf 3.0664 &	\bf 1.3084 &	0.884\\
\hline
\multirowcell{5}{\rotatebox[origin=c]{90}{ETTm1}}
  &	24 &	0.3230 &	0.3690 &	\bf 0.3150 &	0.3886 &	0.3351 &	\bf 0.3875 &	0.975\\
  &	48 &	0.4940 &	0.5030 &	\bf 0.4454 &	\bf 0.4620 &	0.4726 &	0.4702 &	0.902\\
  &	96 &	0.6780 &	0.6140 &	0.4641 &	\bf 0.4823 &	\bf 0.4543 &	0.4831 &	0.670\\
  &	288 &	1.0560 &	0.7860 &	0.6814 &	0.6312 &	\bf 0.6185 &	\bf 0.5991 &	0.586\\
  &	672 &	1.1920 &	0.9260 &	1.1365 &	0.8572 &	\bf 1.1273 &	\bf 0.8412 &	0.946\\
 \hline
\end{tabular}
\caption{Forecasting results for multivariate  time-series.}
\end{table}

\section{BUSINESS PROCESSES}
Having investigated applications of Query Selector algorithms in the area of  time series forecasting related to data points 
gathered by technical devices such as ETT dataset, we extended our research to other areas of time series forecasting.  
One of the most appealing and interesting areas is area related to predictive business process monitoring. 
These challenges  are related to predicting the evolution of ongoing business processes using models built on historical data. 
The very compelling review of more traditional approaches (like: Naive Bayes Classification, Support Vector Regression) 
to this problem can be found in 
\cite{polato2016time}. 
As in so many other cases the deep learning methods and algorithms applied to this field of research turned out to yield 
the better results and using LSTM neural networks   the authors of \cite{Tax_2017} 
achieved the state of the art results on two popular datasets containing data related to business processes.
Using QuerySelector we were able to outperform their results. Precise results are resumed in Table \ref{AccT}.

As structure of business processes is  completely different from that of previous examples, we need to introduce some notation 
which will explain our theoretical setup. We  follow  \cite{polato2016time} and  \cite{Tax_2017} since 
we would like to precisely compare the effectiveness of both methods.

The central notion is {\bf trace} which is a mathematical representation of a case 
(i.e. execution of a specific instance of a business process). Trace is a time-ordered finite sequence of {\bf events}.
Each event can have different properties but for us only its timestamp and type of activity are essential. 
They are given by two functions $\pi_T:\,{\mathcal E}\longrightarrow {\mathcal T}$ and 
$\pi_A:\,{\mathcal E}\longrightarrow {\mathcal A}$ where $\mathcal E$ is the universe of all possible events, 
$\mathcal T$ is time domain and $\mathcal A$ is a finite set of activity types. 

If $\sigma=(\sigma_1, \sigma_2,\dots,\sigma_n)$, $\sigma_i\in\mathcal E$, is a trace, we assume that 
$\sigma_i\neq \sigma_j$ for $i\neq j$, 
and $\pi_T(\sigma_i)\leq \pi_T(\sigma_j)$ for $i<j$ (time-ordering). Functions $\pi_T$ and $\pi_A$ have 
natural extensions to traces (in fact, to any sequences of events): $\pi_T(\sigma)=(\pi_T(\sigma_1),\pi_T(\sigma_2),\dots,\pi_T(\sigma_n))$ 
and similarly for $\pi_A$. A nonempty set of traces is called {\bf event log} and it is a mathematical abstraction of 
dataset.

Our objective is to predict  activity type of future events i.e. to find 
 $\pi_A(\sigma_{k+1})$
given $k\leq n$ first events in a trace $\sigma=(\sigma_1, \sigma_2,\dots,\sigma_n)$.

The two datasets we dealed with are Helpdesk\footnote{\url{https://data.mendeley.com/datasets/39bp3vv62t/1}} and BPI Challenge 
2012\footnote{\url{doi:10.4121/uuid:3926db30-f712-4394-aebc-75976070e91f}}.  
The data contained in those datasets is of the form of  event logs of particular business processes.
Usually, event logs are well structured and each event of a business process is recorded and it refers to an actual 
activity of a particular case. Additionally, it contains other information such as the originator of an activity
or the timestamp.
Since our main goal is to investigate the effectiveness of our Ouery Selector method we follow strictly the setup of 
\cite{Tax_2017} 
in order to be able to compare the results of forecasting algorithms on those datasets. 

The first dataset, Helpdesk contains data logs concerning the ticketing management process of the help desk of an 
Italian software company. In particular, this process consists of
9 activities: it starts with the insertion of a new ticket and then a seriousness level is applied. 
Then the ticket is managed by a resource and it is processed. When the problem is resolved it is closed and 
the process instance ends. This log contains around 3,804 cases and 13,710 events. 

The BPI Challenge 2012 dataset contains real-world event logs from the 2012  BPI challenges (van Dongen 2012). 
The anonymized data from a Dutch financial institute’s loan or overdraft application process was gathered in total of 
262,000 events in 13,087 instances that come from three subprocesses: one that tracks the states of the application, 
another that tracks the states of the offer, and a third that tracks the states of work items associated with the application. 
Events from the third subprocess are classified further into type schedule (work items scheduled for execution), 
type start (work items on which work has started), or type complete (work items on which work has been completed). 
In the context of predicting the coming events and their timestamps we are not interested in events that are performed 
automatically. Thus, we narrow down our evaluation to the work items subprocess, which contains events that are manually 
executed. Further, we alter the log to retain only events of type complete.

\begin{table}\label{AccT}
\begin{center}
\begin{tabular}{|l|c|c|}
\hline
& \multicolumn{2}{c|}{Accuracy}\\
\hline
Dataset & LSTM & QuerrySelector\\
\hline
Helpdesk & 0.7123 &  0.743 \\
BPI'12 & 0.7600 &  0.790 \\
\hline
\end{tabular}
\end{center}
\caption{Comparison of accuracy results. Results for LSTM are the best results reported in \cite{Tax_2017}.}
\end{table}

\noindent{\bf Acknowledgements:}
Special thanks goes to Grzegorz Klimek for help in dealing with datasets.
\break

\bibliographystyle{hapalike}

\bibliography{QuerySelector}

\end{document}